\documentclass[a4paper,12pt]{article}
\usepackage{indentfirst} 
\usepackage{url}         
\usepackage{upquote}     
\usepackage{xcolor}      
\usepackage{orcidlink}   
\usepackage{fontawesome5}

\definecolor{githubgray}{HTML}{333333}
\newcommand{\githublink}[1]{
  \href{https://github.com/#1}{
    \textcolor{githubgray}{\faGithub}
    \hspace{0.5em}
    \texttt{#1}
  }
}

\usepackage{etoolbox}                       
\AtBeginDocument{\hypersetup{hidelinks}}    

\usepackage{authblk}                        

\usepackage{geometry}			
\usepackage{amsmath,amsfonts,amssymb,amsthm,mathtools}

\usepackage{graphicx}					
\graphicspath{{figures/}}		        
\usepackage{caption}                    
\usepackage{array,tabularx,tabulary,booktabs}	

\usepackage{cite}						

\usepackage{nomencl}                    
\begin{document}

\title{A Comparative Benchmark of Large Language Models \\ for Labelling Wind Turbine Maintenance Logs}
\author[1]{Max Malyi\orcidlink{0000-0002-1503-9798}\thanks{Corresponding author: \url{Max.Malyi@ed.ac.uk}}}
\author[1]{Jonathan Shek\orcidlink{0000-0001-5734-2907}}
\author[1]{Alasdair McDonald\orcidlink{0000-0002-2238-3589}}
\author[2]{André Biscaya\orcidlink{0000-0002-8158-4284}}

\affil[1]{Institute for Energy Systems, School of Engineering, The University of Edinburgh, Edinburgh, UK}
\affil[2]{Nadara, Lisbon, Portugal}

\date{August, 2025}

\maketitle

\begin{abstract}
Effective Operation and Maintenance~(O\&M) is critical to reducing the Levelised Cost of Energy~(LCOE) from wind power, yet the unstructured, free-text nature of turbine maintenance logs presents a significant barrier to automated analysis. Our paper addresses this by presenting a novel and reproducible framework for benchmarking Large Language Models~(LLMs) on the task of classifying these complex industrial records. To promote transparency and encourage further research, this framework has been made publicly available as an open-source tool. We systematically evaluate a diverse suite of state-of-the-art proprietary and open-source LLMs, providing a foundational assessment of their trade-offs in reliability, operational efficiency, and model calibration. Our results quantify a clear performance hierarchy, identifying top models that exhibit high alignment with a benchmark standard and trustworthy, well-calibrated confidence scores. We also demonstrate that classification performance is highly dependent on the task's semantic ambiguity, with all models showing higher consensus on objective component identification than on interpretive maintenance actions. Given that no model achieves perfect accuracy and that calibration varies dramatically, we conclude that the most effective and responsible near-term application is a Human-in-the-Loop system, where LLMs act as a powerful assistant to accelerate and standardise data labelling for human experts, thereby enhancing O\&M data quality and downstream reliability analysis.

\vspace{0.25cm}
\noindent\rule{\linewidth}{0.3pt}
\vspace{0.25cm}

data labelling, text classification.

\vspace{0.25cm}
\noindent\rule{\linewidth}{0.3pt}
\vspace{0.25cm}

\noindent The project source code is hosted in an open source GitHub Repository:

\vspace{0.25cm}

\noindent \githublink{mvmalyi/wind-farm-maintenance-logs-labelling-with-llms}

\end{abstract}

\newpage

\tableofcontents

\makenomenclature
\nomenclature{AI}{Artificial Intelligence}
\nomenclature{API}{Application Programming Interface}
\nomenclature{KPI}{Key Performance Indicator}
\nomenclature{LCOE}{Levelised Cost of Energy}
\nomenclature{LLM}{Large Language Model}
\nomenclature{O\&M}{Operation and Maintenance}
\nomenclature{SCADA}{Supervisory Control and Data Acquisition}
\printnomenclature[2 cm]

\newpage

\section{Introduction}
\label{sec:introduction}

Effective operation and maintenance~(O\&M) strategies are paramount to reducing the Levelised Cost of Energy~(LCOE) for wind farms, with O\&M expenditures constituting up to a third of a wind turbine's lifetime costs. A prerequisite for optimising these strategies and extending turbine lifetimes is robust reliability analysis, which depends entirely on the quality and structure of the underlying maintenance data. Central to this is the wealth of information contained within maintenance logs, often referred to as work orders, which document all interventions performed on the turbines. However, the wind industry currently lacks a uniform methodology for collecting and analysing this critical data, which is frequently recorded as unstructured free text characterised by inconsistent terminology, abbreviations, and a lack of standardised formatting \cite{hodkiewicz_cleaning_2016, hahn_recommended_2017}. The consequences of this unprocessed data are not trivial, foundational studies have shown that the choice of data cleaning methodology can alter key reliability parameters, such as a component's characteristic life, by as much as a factor of three \cite{hodkiewicz_cleaning_2016}. This absence of structure presents a significant barrier to automated analysis, compelling operators to rely on time-consuming and costly manual data processing to extract actionable insights for reliability analysis and performance monitoring \cite{salo_work_2019}.

These maintenance logs are typically generated by technicians and stored in Computerised Maintenance Management Systems, such as IBM Maximo or SAP. While indispensable, these records are primarily designed for operational tracking rather than analytical purposes. Consequently, the descriptive fields often contain a mix of technical jargon, site-specific abbreviations, and sometimes typographical errors, making the raw text a rich but challenging source of information. This lack of standardisation means that a significant portion of the potential value from this data remains untapped, as it cannot be easily aggregated or analysed at scale.

Historically, efforts to digitise and structure this maintenance information have employed traditional machine learning and text classification techniques. Early digitalisation workflows, for instance, demonstrated the feasibility of automated structuring by using methods like Support Vector Machines to classify maintenance activities according to domain-specific taxonomies \cite{lutz_digitalization_2022}. Such automation was highly motivated by the immense effort required by purely manual approaches, where processing a single wind farm's dataset could take a specialist two working weeks \cite{salo_analysis_2017}. Subsequent studies have compared automated text classification against manual expert labelling and Artificial Intelligence~(AI) assisted tagging, confirming that while automation drastically reduces processing time, the accuracy of Key Performance Indicator~(KPI) extraction remains a critical challenge \cite{lutz_kpi_2023}. However, recent investigations have revealed a fundamental uncertainty when using manual processing as the gold standard. When the same dataset was labelled independently by two different expert organisations, the resulting failure rate KPIs for some subsystems differed by over 300\% \cite{walgern_impact_2024}, highlighting the inherent subjectivity and inconsistency of manual analysis.

The recent advent of Large Language Models~(LLMs) represents a paradigm shift in natural language processing, offering unprecedented capabilities in understanding context, semantics, and domain-specific jargon. Unlike traditional machine learning models that require extensive feature engineering and large labelled datasets, modern LLMs can perform complex classification and information extraction tasks in a zero-shot or few-shot setting. However, their application in industrial settings is often constrained by data privacy and the high computational cost of proprietary models, motivating the exploration of locally-hosted, open-source alternatives \cite{walshe_automatic_2025}. The potential for these models in the wind energy sector is substantial, yet still nascent. Recent work by Walgern~et~al.~(2024) explicitly identifies next-generation LLMs, such as GPT-4 (relevant at the time of that study), as the future frontier for standardising maintenance data, suggesting that fine-tuning on domain-specific datasets could significantly enhance classification performance and overcome the limitations of current classifiers \cite{walgern_impact_2024}. Furthermore, the application of LLMs is expanding beyond mere classification towards predictive and advisory roles. Pioneering research by Walker~et~al.~(2024) explores the development of specialised LLM-based conversational agents capable of recommending repair actions from sequences of Supervisory Control and Data Acquisition~(SCADA) alarms \cite{walker_safellm_2024}. This work also confronts the inherent challenges of LLMs, such as hallucinations (i.e., the generation of plausible but factually incorrect or nonsensical text) and unsafe responses, proposing domain-specific safety monitoring to ensure their reliability in safety-critical industrial environments.

Despite this promising trajectory, a critical gap remains in the literature. While foundational studies have focused on traditional text classifiers \cite{walgern_impact_2024} and the separate challenge of record linkage between data silos \cite{cox_enrichment_2022}, the practical application of modern Large Language Models to this domain remains under-explored. The literature lacks a comparative benchmark that assesses the performance, cost, and reliability of current state-of-the-art LLMs for this task. Consequently, a comprehensive, comparative benchmarking of different state-of-the-art LLMs for these more complex, real-world tasks has not yet been performed.

To address this gap, this paper introduces a novel, open-source framework designed to systematically benchmark LLMs for the task of classifying unstructured wind turbine maintenance logs. The primary contribution of this work is therefore the methodology and the reusable tool itself, which provides the industry with a standardised means of evaluating and deploying LLMs for this critical data-structuring task. We demonstrate this framework’s utility by conducting a comprehensive benchmark of leading proprietary and open-source LLMs, evaluating their performance on alignment, cost, and the quality of their confidence scores. This study provides a foundational assessment to support the adoption of LLMs in wind energy O\&M and encourages further research into their integration with existing reliability analysis workflows.

\section{Methods}
\label{sec:methods}

This study employs a detailed and reproducible methodological framework to benchmark a diverse set of LLMs for the task of classifying unstructured maintenance logs from wind turbines. The methodology is designed to provide a comprehensive evaluation of the models based on their accuracy, efficiency, and cost-effectiveness, while also transparently addressing the inherent limitations of the experimental setup. The process encompasses three primary stages: data preparation, the LLM benchmarking workflow with a description of the classification task and novel aspects of the prompt engineering, and the definition of evaluation metrics.

\subsection{Data Preparation}
The dataset was prepared through a multi-stage curation pipeline to ensure high informational quality. Initially, the data was scoped to turbine-specific logs and component nomenclature was standardised. Rule-based cleaning, a common practice for such industrial data \cite{hodkiewicz_cleaning_2016, salo_analysis_2017}, was then applied to remove syntactic noise and textual redundancies based on Levenshtein distance. To enhance diversity, the dataset was balanced by down-sampling the majority class and applying cascading frequency caps on duplicate entries. The final stage involved semantic de-duplication: logs were converted into Sentence-BERT embeddings (a model for generating semantically meaningful sentence vectors) and clustered using the HDBSCAN algorithm (a density-based clustering method), implemented with Python's scientific computing libraries. Only a small, representative sample of logs was retained from each semantic cluster, resulting in a final dataset of more than 400 information-rich logs designed for robust LLM evaluation. A sample of the data format in English translation is provided in Table~\ref{tab:data_format_sample}.

\begin{table}[!ht]
\centering
\caption{A sample of the curated maintenance logs, demonstrating the data format provided to the LLMs.}
\label{tab:data_format_sample}
\resizebox{\textwidth}{!}{%
\begin{tabular}{@{}llp{7cm}p{7cm}@{}}
\toprule
\textbf{Component Code} & \textbf{Component Name} & \textbf{Log Description} & \textbf{Additional Observations} \\
\midrule
\texttt{MDA10} & Rotor Blades & Inspecting the damage on WTG05 & We found that we actually have two damages: new blade damage discovered \\
\addlinespace
\texttt{MDX10} & Central Hydraulic System & Stops with the error & Sludge pitch hydraulic station \\
\bottomrule
\end{tabular}%
}
\end{table}

Given that the smaller, open-source models are predominantly trained on English-language corpora, the input logs for these models were translated to English prior to the benchmark using the Gemini-2.5-Pro model to ensure a high-quality, context-aware translation. The larger, proprietary Application Programming Interface~(API) based models were evaluated directly on the original (Portuguese) text to leverage their native multilingual capabilities.

\subsection{LLM Benchmarking Approach}
The core of the methodology is a systematic benchmarking process designed to test a range of contemporary LLMs, spanning proprietary APIs and locally-hosted open-source models.

\subsubsection{Model Selection}
The models were selected to represent the state-of-the-art landscape in mid-2025, comprising the latest model families from key industry providers (OpenAI and Google) to whom API access was available, and a suite of leading open-source models from major developers (such as Microsoft, Google, Meta, and Mistral, respectively) chosen for their strong performance-to-size ratio and ability to run on the available local hardware.

\vspace{0.5cm}

\noindent\textbf{Proprietary API-based Models:}
\begin{itemize}
    \item OpenAI: GPT-5, GPT-5 Mini, GPT-5 Nano, GPT-o3, and GPT-o4 Mini.
    \item Google: Gemini-2.5 Pro and Gemini-2.5 Flash.
\end{itemize}

\noindent\textbf{Locally-hosted Open-Source Models:} A suite of models was run locally using the Ollama framework on a MacBook Air with M4 processor and 24 GB RAM. The selected models were chosen based on their ability to run efficiently on available hardware: Phi-4:14b, Gemma-3:12b, Llama-3.1:8b, and Mistral:7b, where the -b suffix denotes the number of model parameters in billions.

\subsubsection{Classification Task and Prompt Engineering}
The primary task for the LLMs was to classify each maintenance log against two predefined, hierarchical taxonomies: \textit{Maintenance Type} and \textit{Issue Category}. The label schemas for these categories were developed through a hybrid approach. An initial list of common maintenance and issue types was formulated based on a review of existing literature and manual observation of the data. This list was then refined and validated by providing a comprehensive list of unique log descriptors extracted for each component to a reasoning LLM, which then provided suggestions for modifications in the labels list.

The final schema consisted of 16 \textit{Maintenance Type}  and 26 \textit{Issue Category} labels. A key innovation in this methodology was the dynamic filtering of prompts. For each log entry, the system first identified the component code and then retrieved only the subset of \textit{Maintenance Type} and \textit{Issue Category} labels relevant to that specific component. This filtered list was then used to construct the prompt (i.e., the set of instructions and context provided to the model), a technique designed to improve model accuracy by reducing contextual noise and the risk of model hallucinations, while also decreasing token consumption (the basic units of text that the model processes). This approach is motivated by findings that LLMs can exhibit performance degradation on high-cardinality tasks when presented with an overly large label space in a single prompt \cite{walshe_automatic_2025}. The complete lists of used maintenance types and issue categories, along with the component code legend, are available in the project's public repository.

\subsubsection{Framework Enhancements for Robustness and Reproducibility}
To enhance the rigour of the study, the methodological framework was substantially refined. The experimental workflow was re-implemented using a modular, object-oriented code structure with distinct Python classes for handling different model APIs, and separated into two distinct stages: a benchmark execution stage for generating raw data, and an in-depth analysis stage for processing results. To ensure full traceability and prevent data loss, every experimental run automatically archives all outputs, including logs and results, into a unique, timestamped directory. All key parameters, including model definitions, token pricing, and prompt structures, were externalised to central configuration files to facilitate rapid and transparent iteration.

A key innovation was introduced in the prompt engineering strategy through a dynamic mapping system. This supports both inclusion and exclusion rules, allowing for more flexible and concise definitions of the label space presented to each model. The prompt template was also refined with explicit instructions on brevity and output structure to improve parsing reliability and mitigate API errors, such as exceeding maximum token limits.

Finally, the evaluation process was extended beyond standard classification metrics. To analyse model calibration as a model's awareness of its own certainty, a procedure was introduced to correlate the models' self-reported confidence scores with their F1-scores against the labels set to be ground truth. The concept of ground truth labels was also made flexible, enabling analyses of not only accuracy (against manually verified labels, a methodology used in foundational data fusion studies \cite{cox_enrichment_2022}) but also agreement (against a model consensus) and alignment (against the output of a designated benchmark model), though this preliminary study focuses primarily on the latter two. This multi-faceted approach was chosen to mitigate the known challenges of expert label subjectivity, where different experts can produce significantly different reliability KPIs from the same dataset \cite{walgern_impact_2024}.

\subsubsection{Other Noteworthy Methodological Decisions}
Beyond the core framework design, several other decisions were made to enhance the utility and robustness of the benchmark:
\begin{itemize}
    \item \textbf{Rigorous Output Validation:} The framework was designed for comprehensive error logging that went beyond capturing technical API failures. A strict JSON data schema was enforced on the LLM outputs to validate their structure upon receipt. Furthermore, any outputs identified as obvious hallucinations or nonsensical responses during the manual review process were systematically logged, providing a richer basis for the error rate analysis.
    \item \textbf{Exploratory Data Extraction:} The prompt also instructed models to populate an optional \textit{Specific Issue} field with a more granular, human-readable summary of the problem if one could be inferred from the text. While this feature was included to test the models' summarisation capabilities for potential future applications, a quantitative analysis of the accuracy of these generated descriptions was beyond the scope of this study.
\end{itemize}

\subsubsection{Methodological Limitations}
It is important to note several limitations inherent to this study's scope:
\begin{itemize}
    \item The curated dataset, while high in quality, represents only the most informative segment of real-world maintenance logs, many of which lack sufficient semantic value for classification.
    \item The selection of locally-hosted models was constrained by the available computational hardware, precluding the use of more resource-intensive models.
    \item The evaluation focuses on model alignment with a top-performing benchmark and agreement with a model consensus. While the framework is equipped to measure accuracy against expert-verified labels, such an analysis was beyond the scope of this project.
\end{itemize}

\subsection{Evaluation Metrics}
The performance of each LLM was assessed against a comprehensive set of quantitative and qualitative criteria to provide a holistic view of its suitability for this industrial application.
\begin{itemize}
    \item \textbf{Accuracy, Precision, Recall, and F1-Score:} The correctness of the assigned labels was evaluated using a suite of standard classification metrics, calculated from a manual check of a random sample of the outputs. In this context, precision measures the proportion of retrieved labels that are truly correct (i.e., how many of the labels the model assigned are right), recall measures the proportion of all correct labels in the dataset that the model successfully identified (i.e., how many of the right labels did the model find), and the F1-score provides a single metric that balances both. These are defined as:
    \begin{gather}
        \text{Acc} = \frac{TP + TN}{TP + TN + FP + FN} \\
        P = \frac{TP}{TP + FP} \\
        R = \frac{TP}{TP + FN} \\
        F_1 = 2 \cdot \frac{P \cdot R}{P + R}
    \end{gather}
    where $P$ is Precision, $R$ is Recall, and $TP$, $TN$, $FP$, and $FN$ are the respective counts of true positives, true negatives, false positives, and false negatives, evaluated against the benchmark labels.

    \item \textbf{Inter-Model Agreement:} To measure the consistency between models, a consensus-based agreement score was calculated. For each log, the consensus label ($L_{C,i}$) was established by finding the statistical mode of the labels assigned by the entire set of models. The agreement score ($Agr_m$) for an individual model $m$ was then defined as the proportion of its labels that matched the consensus:
    \begin{equation}
        Agr_m = \frac{1}{N} \sum_{i=1}^{N} \mathbb{I}(L_{i,m} = L_{\text{con},i})
    \end{equation}
    where $N$ is the total number of logs, $L_{i,m}$ is the label from model $m$ for log $i$, and $\mathbb{I}$ is the indicator function, which is 1 if the condition is true and 0 otherwise.

    \item \textbf{Token Usage and Cost:} For the API-based models, the total financial cost, $C_{\text{total}}$, was calculated based on the token consumption for the entire dataset of $N$ logs. This is formalised as:
    \begin{equation}
        C_{\text{total}} = \sum_{i=1}^{N} (T_{\text{in},i} \cdot P_{\text{in}} + T_{\text{out},i} \cdot P_{\text{out}})
    \end{equation}
    where $T_{\text{in},i}$ and $T_{\text{out},i}$ are the number of input and output tokens for the $i$-th log, and $P_{\text{in}}$ and $P_{\text{out}}$ are the respective costs per token for the given model.

    \item \textbf{Processing Speed:} The computational efficiency was quantified as throughput, $S$, measured in logs processed per second:
    \begin{equation}
        S = \frac{N}{t_{\text{total}}}
    \end{equation}
    where $N$ is the total number of logs in the dataset and $t_{\text{total}}$ is the total execution time in seconds.

    \item \textbf{Error and Hallucination Rate:} The reliability of each model was measured by a combined error rate, $E_{H}$, which includes both technical failures and observed content issues:
    \begin{equation}
        E_{H} = \frac{N_{\text{fail}} + N_{\text{hall}}}{N_{\text{total}}}
    \end{equation}
    where $N_{\text{fail}}$ is the count of failed script requests, $N_{\text{hall}}$ is the count of outputs identified as obvious hallucinations during manual review, and $N_{\text{total}}$ is the total number of logs processed.

    \item \textbf{Confidence Score:} Where available from the model's output, a self-reported certainty score for each classification was collected as a supplementary, qualitative metric.
    
\end{itemize}

\subsection{Source Code Availability}
To ensure the reproducibility and transparency of this research, the complete source code, configuration files, and supplementary materials have been made publicly available. The open-source repository includes the modular Python scripts for benchmark execution and analysis, the full label schemas, component code legends, and the dynamic prompt templates used in this study. The repository can be accessed on our \textcolor{blue}{\href{https://github.com/mvmalyi/wind-farm-maintenance-logs-labelling-with-llms}{GitHub Repository}}. 

\section{Results}
\label{sec:results}

The performance of the LLMs was evaluated across four key domains: operational efficiency and reliability, alignment with a benchmark model, model calibration, and inter-model agreement. A comprehensive summary of the key performance and alignment metrics is presented in Table~\ref{tab:master_summary}.

\begin{table}[!ht]
\centering
\caption{Overall Performance and Alignment Summary of All Tested Models.}
\label{tab:master_summary}
\resizebox{\textwidth}{!}{%
\begin{tabular}{lcccccc}
\hline
\textbf{Model} & \textbf{Throughput (logs/s)} & \textbf{Total Tokens} & \textbf{Est. Cost (\$)} & \textbf{Error Rate (\%)} & \textbf{Average F1 Score} & \textbf{Average Consensus} \\
\hline
\texttt{gpt-5} & 0.10 & 341,596 & 2.31 & 0.00 & \textcolor{gray}{1.00} & 0.80 \\
\texttt{gpt-5-mini} & 0.13 & 277,334 & 0.33 & 0.00 & 0.74 & 0.91 \\
\texttt{gpt-5-nano} & 0.12 & 520,277 & 0.16 & 0.00 & 0.73 & 0.87 \\
\texttt{gpt-o3} & 0.16 & 230,481 & 1.09 & 0.00 & 0.83 & 0.86 \\
\texttt{gpt-o4-mini} & 0.17 & 269,000 & 0.77 & 0.00 & 0.74 & 0.87 \\
\texttt{gemini-2.5-pro} & 0.11 & 148,351 & 0.37 & 0.00 & 0.80 & 0.89 \\
\texttt{gemini-2.5-flash} & 0.18 & 145,325 & 0.09 & \textcolor{red}{2.06} & 0.74 & 0.88 \\
\texttt{phi4\_14b} & 0.09 & 144,550 & 0.00 & \textcolor{red}{0.26} & 0.63 & 0.85 \\
\texttt{gemma3\_12b} & 0.12 & 145,136 & 0.00 & 0.00 & 0.64 & 0.86 \\
\texttt{llama3.1\_8b} & 0.16 & 144,083 & 0.00 & 0.00 & 0.62 & 0.83 \\
\texttt{mistral\_7b} & 0.14 & 178,205 & 0.00 & \textcolor{red}{1.29} & 0.62 & 0.76 \\
\hline
\end{tabular}
}
\end{table}

\subsection{Operational Efficiency and Reliability}
\begin{figure}[h!]
    \centering
    \includegraphics[width=\textwidth]{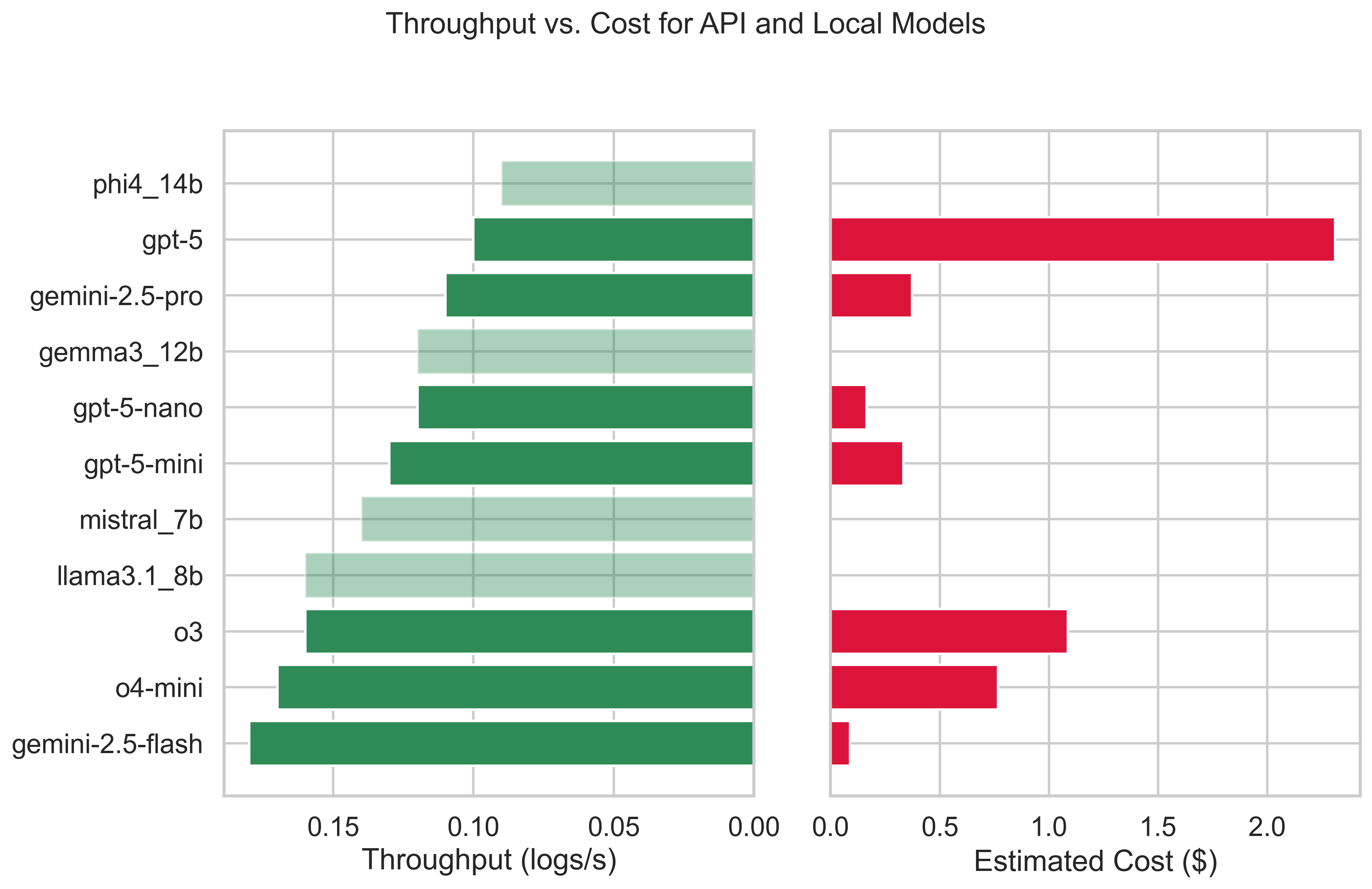}
    \caption{Throughput (logs/s) vs. Estimated Cost (\$) for API and local models. The left panel shows processing speed (higher is better), with local models in a lighter shade. The right panel shows the financial cost for API models (lower is better).}
    \label{fig:throughput_cost}
\end{figure}

The benchmark revealed a distinct trade-off between processing speed, cost, and reliability. As shown in Figure~\ref{fig:throughput_cost}, \texttt{gemini-2.5-flash} was the fastest and most cost-effective API model, but this was offset by the technical error rate of 2.06\%. In contrast, running \texttt{gpt-5} was perfectly reliable but with the highest cost and slowest throughput. The throughput of open-source models was competitive but is contingent on the local hardware used in this study. Reliability issues in open-source models were primarily content-based, with \texttt{mistral\_7b} exhibiting the highest rate of hallucinated labels, causing concerns in the trustworthiness of the whole output.

\subsection{Classification Alignment}
\begin{figure}[h!]
    \centering
    \includegraphics[width=\textwidth]{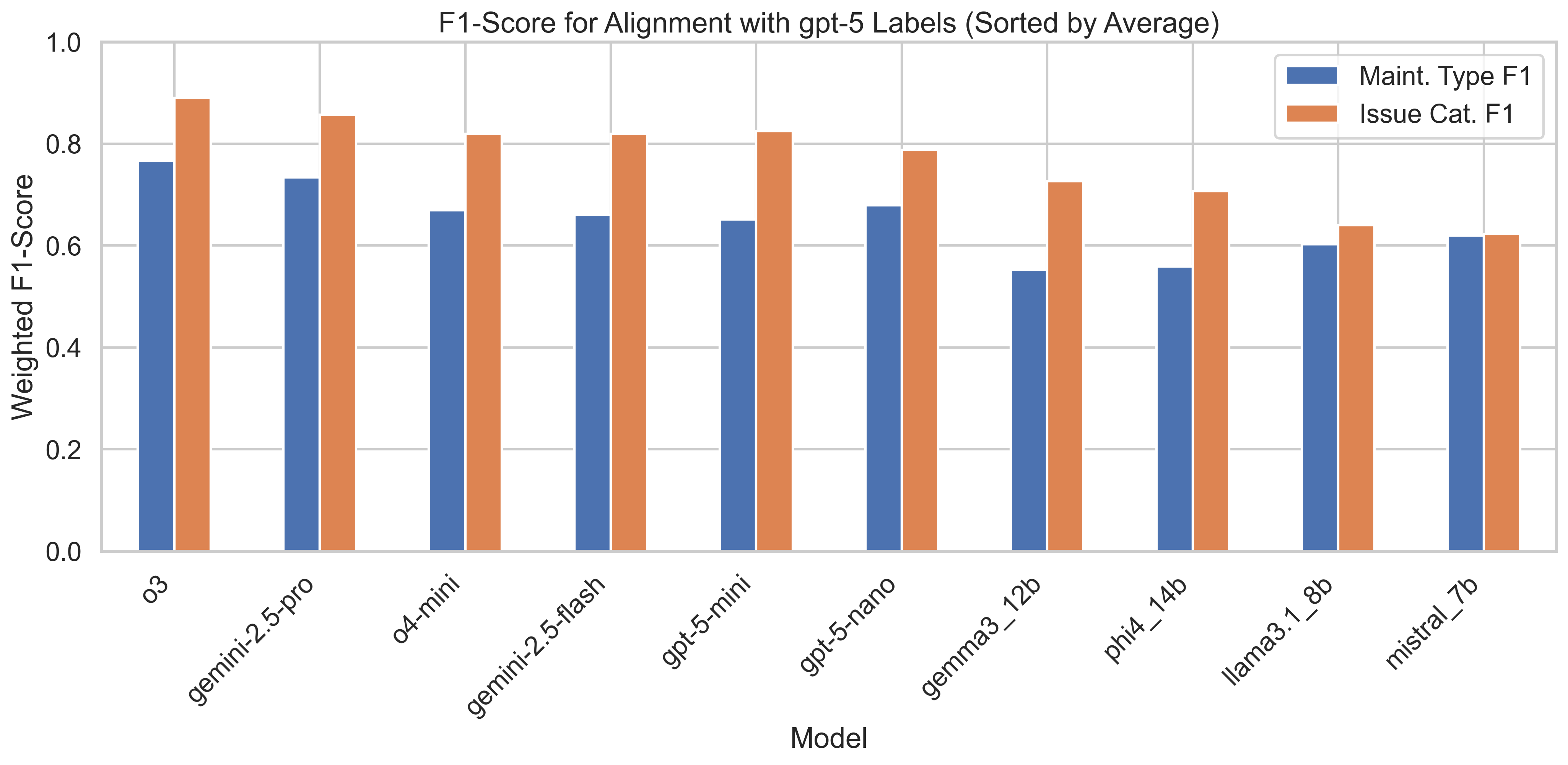}
    \caption{Weighted F1-Scores for each model's alignment with the \texttt{gpt-5} benchmark, sorted by average performance across both tasks.}
    \label{fig:f1_scores}
\end{figure}

Using \texttt{gpt-5} as the benchmark, a clear hierarchy of alignment emerged (Figure~\ref{fig:f1_scores}). The \texttt{gpt-o3} and \texttt{gemini-2.5-pro} models demonstrated the highest alignment, achieving average F1-scores of 0.83 and 0.80, respectively. A mid-tier of models, including the smaller OpenAI and Gemini variants, performed competently, while the open-source models consistently showed the most divergent logic.

\begin{figure}[h!]
    \centering
    \includegraphics[width=0.8\textwidth]{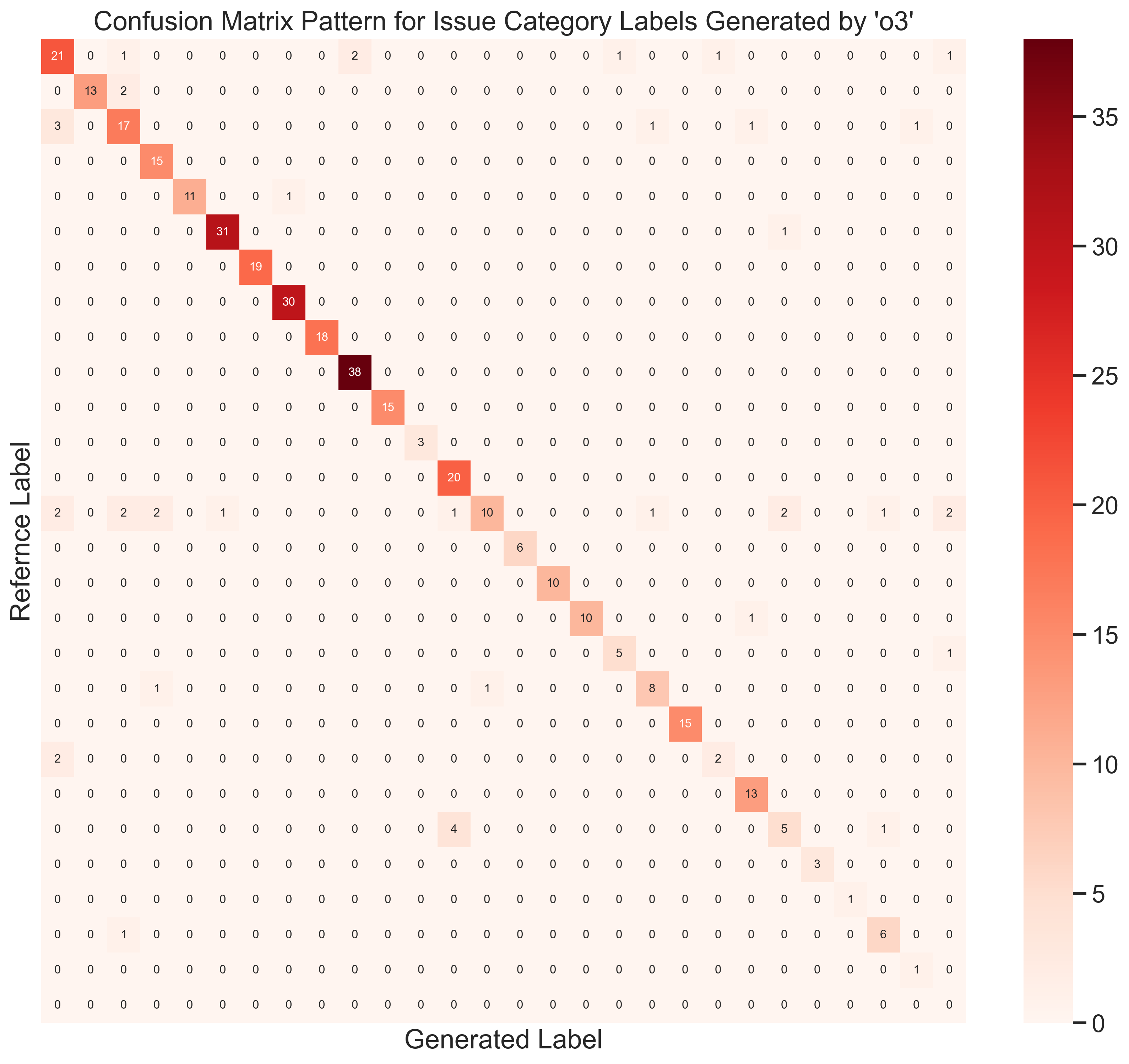}
    \caption{Confusion matrix for the \textit{Issue Category} labels generated by \texttt{gpt-o3} against the \texttt{gpt-5} reference. The clear diagonal pattern indicates high agreement.}
    \label{fig:confusion_matrix}
\end{figure}

A critical finding was the universal difference in performance between the two classification tasks. Across all models, alignment was significantly higher for the more objective \textit{Issue Category} task than for the more interpretive \textit{Maintenance Type} task. For instance, the top-performing \texttt{gpt-o3} model achieved an F1-score of 0.89 on \textit{Issue Category} classification but only 0.77 on \textit{Maintenance Type}, quantifying the increased difficulty of the latter. The confusion matrix for \texttt{gpt-o3}'s \textit{Issue Category} predictions (Figure~\ref{fig:confusion_matrix}) shows a strong diagonal, indicating high agreement with the benchmark on most component-level identifications.

\subsection{Model Calibration and Inter-Model Agreement}
\begin{figure}[h!]
    \centering
    \includegraphics[width=0.9\textwidth]{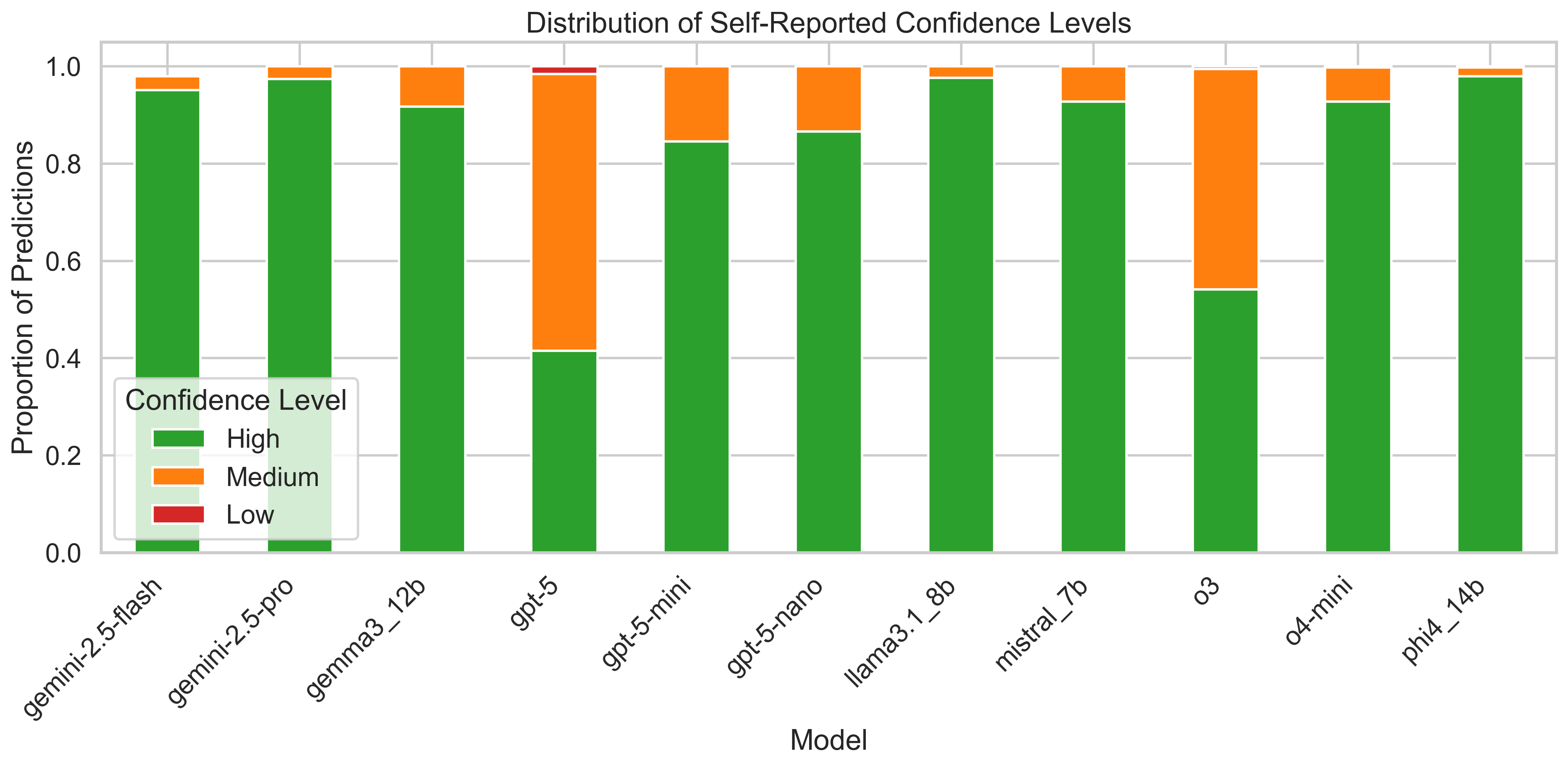}
    \caption{Distribution of self-reported confidence levels across all models.}
    \label{fig:model_confidence_distribution}
\end{figure}

\begin{figure}[h!]
    \centering
    \includegraphics[width=\textwidth]{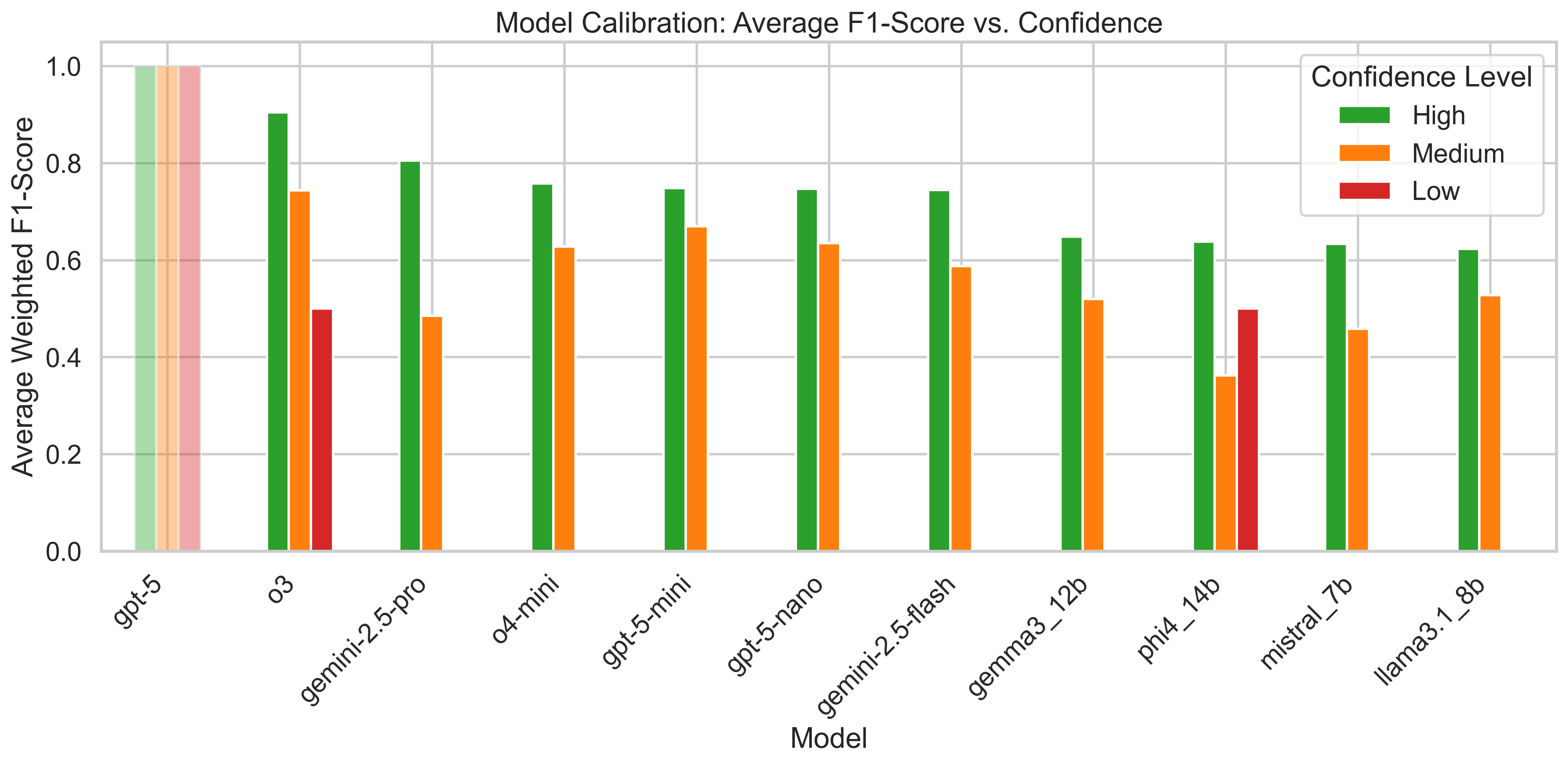}
    \caption{Model Calibration: Average F1-Score vs. Self-Reported Confidence Level. \texttt{gpt-5} is shown as the benchmark (dimmed bars).}
    \label{fig:calibration}
\end{figure}

Model calibration as the model's awareness of its own certainty proved to be a critical differentiator. The distribution of these self-reported confidence levels across the models is visually presented in Figure \ref{fig:model_confidence_distribution}, offering further insight into their calibration tendencies. As shown in Figure~\ref{fig:calibration}, \texttt{gpt-o3} exhibited excellent calibration. Its labels assigned with 'High' confidence achieved an average F1-score of 0.91, whereas its 'Low' confidence labels scored just 0.50. This strong positive correlation contrasts sharply with poorly calibrated models like \texttt{phi4\_14b}, which assigned 'High' confidence to labels that only achieved an F1-score of 0.64, demonstrating significant overconfidence.

\begin{figure}[h!]
    \centering
    \includegraphics[width=0.8\textwidth]{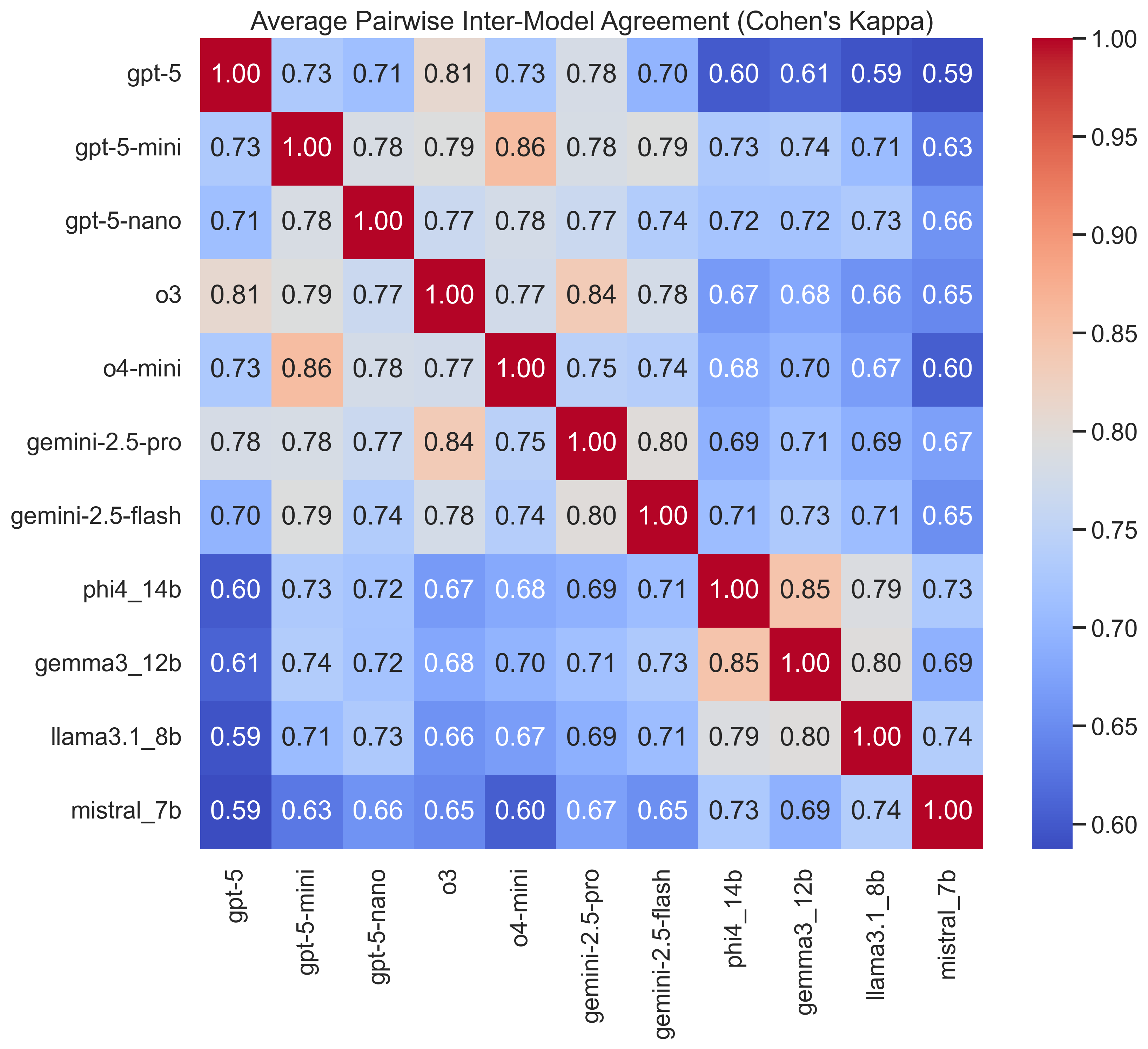}
    \caption{Average Pairwise Inter-Model Agreement (Cohen's Kappa). Warmer colours indicate higher agreement between models.}
    \label{fig:kappa_heatmap}
\end{figure}

The average pairwise inter-model agreement, measured by Cohen's Kappa, is visualised in Figure~\ref{fig:kappa_heatmap}. A Kappa score above 0.81 indicates high agreement, while scores between 0.61 and 0.80 are considered substantial. The heat-map confirms a central cluster of high agreement between the proprietary models from OpenAI and Google. The open-source models, particularly \texttt{mistral\_7b}, show the most distinct reasoning patterns, with lower Kappa scores relative to this central cluster, which may be related to potential hallucinations.

\section{Discussion}
\label{sec:discussion}

A key takeaway from this study is that modern LLMs are exceptionally well-suited for this industrial classification task, representing a significant leap over both traditional natural language processing methods and purely manual processing. Unlike previous techniques that rely on extensive and often brittle feature engineering to handle domain-specific jargon, abbreviations, and misspellings, LLMs can interpret the context and semantics of free text with minimal pre-processing. This inherent contextual understanding allows them to generalise far more effectively than rule-based or early machine learning models. Furthermore, when compared to manual labelling, LLMs offer unparalleled scalability and consistency. While manual processing is susceptible to human subjectivity and fatigue, a factor that contributes to significant inter-expert discrepancies \cite{walgern_impact_2024}, an LLM applies the same logic to every single log, enabling the rapid and uniform processing of vast historical datasets that would be infeasible to label by hand.

The results of this benchmark confirm that while LLMs are powerful tools for industrial data labelling, they are not a monolithic, plug-and-play solution. The observed trade-off between efficiency and reliability highlights that the optimal model choice is application-specific. The higher hallucination rate in smaller open-source models, likely due to a weaker adherence to complex prompt constraints, underscores the findings of Lutz~et~al.~(2023), where simpler models struggled with domain-specific data  \cite{lutz_kpi_2023}. Furthermore, a significant advantage of leveraging large, proprietary models is their inherent multilingual capability. By processing logs in their original language, they obviate the need for a pre-translation step, which not only saves effort but also eliminates a potential source of inaccuracy or lost nuance that can arise from translation when there is a language barrier between maintenance personnel and data analysts.

The benchmark also illuminates the critical trade-offs between proprietary and open-source models for an industrial use case. Proprietary models, accessed via APIs, offer the significant advantages of ease of use and immediate access to state-of-the-art performance without any upfront hardware investment. However, this convenience comes at the cost of data privacy, as sensitive operational data must be sent to third-party servers, and can lead to vendor lock-in. Conversely, open-source models provide complete control over data and infrastructure, ensuring privacy and allowing for deep customisation. This control, however, requires a substantial investment in computational hardware and the in-house expertise to deploy, fine-tune, and maintain the models. The possibility of fine-tuning is also an important consideration for both approaches. While many proprietary models now offer API-based fine-tuning, this incurs its own cost structure for training and usage, which must be weighed against the hardware and development costs associated with fine-tuning an open-source model locally.

A crucial finding, quantified in the results, is the differing semantic ambiguity of the two classification tasks. The high inter-model agreement and superior F1-scores on the \textit{Issue Category} task suggest it is a relatively objective exercise of identifying physical components. However, the significant divergence and lower F1-scores on the \textit{Maintenance Type} task reveal its interpretive nature. The line between a "repair," a "replacement," or an "inspection" can be blurry, depending on the technician's description. This confirms the challenges noted by Walgern~er~al.~(2024), who found that even human experts disagree substantially on such classifications, questioning the very notion of a single established ground truth and highlighting the inherent uncertainty in manual labelling processes \cite{walgern_impact_2024}.

The analysis of model calibration is perhaps the most critical finding for industrial application. Most models exhibited a high degree of confidence, even when their alignment was moderate. As quantified by the results for \texttt{phi4\_14b}, this tendency towards overconfidence is a significant operational risk. A well-calibrated model like \texttt{gpt-o3}, whose self-reported confidence is a reliable indicator of its actual performance, is substantially more valuable than a poorly calibrated one whose confidence is misleading. This directly addresses the need for trustworthy AI in safety-critical environments. Such reliability is a prerequisite for the development of advanced advisory frameworks like SafeLLM, where the system must be able to gauge the LLM's certainty to decide whether to trust a recommendation or escalate it for human review \cite{walker_safellm_2024}.

Ultimately, for an application of this nature, classification accuracy must be considered the primary driver of model selection. For a large-scale wind farm operator, the computational costs highlighted in this study, a matter of a few dollars to process hundreds of records, are negligible. This minor expenditure is dwarfed by the immense downstream value unlocked by more accurate and trustworthy data. Such data is the bedrock of reliable KPI calculation, which in turn informs multi-million-pound decisions regarding O\&M strategies, spare parts inventory, and predictive maintenance scheduling. The cost of inaccurate data, which can lead to flawed reliability models and suboptimal operational decisions, poses a far greater financial risk than the marginal cost of using a superior LLM. Therefore, prioritising the most accurate and well-calibrated model is not an expense, but a crucial investment in operational intelligence.

While the potential of LLMs as a technology is compelling, our observations on imperfect accuracy and variable calibration lead us to infer that a fully autonomous labelling system is not yet advisable. Instead, we strongly recommend a Human-in-the-Loop system as the most responsible and effective near-term implementation of this technology. Such a system leverages the LLM as a powerful assistant and offers a dual benefit: it can be used both for real-time data standardisation and for labelling large historical archives. In a practical workflow, an LLM can analyse free text at the point of data entry and suggest labels for immediate verification by a technician, or it can process entire legacy datasets, allowing a subject matter expert to efficiently validate or correct the suggestions in batches. This approach offers three key advantages: (1) it dramatically reduces the manual effort and cognitive load of data entry and validation; (2) it improves data standardisation by providing consistent, high-quality suggestions across both new and old data; and (3) it maintains near-perfect accuracy by keeping the human expert as the final arbiter.

\section{Conclusions and Future Work}
\label{sec:conclusion}

This paper presented a novel, open-source framework for benchmarking LLMs on the task of classifying unstructured wind turbine maintenance logs. Our systematic evaluation of leading proprietary and open-source models yielded four key conclusions. First, there is a distinct and quantifiable trade-off between model performance, financial cost, and operational reliability, with no single model excelling across all metrics. Second, model calibration, the reliability of a model's self-reported confidence, is a critical performance differentiator for industrial applications, as many models exhibit significant overconfidence. Third, classification performance is intrinsically linked to the semantic ambiguity of the task. All models demonstrated substantially higher alignment on objective component identification than on the more interpretive classification of maintenance actions. Finally, given that no model achieves perfect accuracy and their reliability varies, a fully autonomous system is not yet advisable, while the Human-in-the-Loop approach with an AI-assisted interface utilising LLMs for data labelling appears feasible.

While study provides a foundational benchmark for applying LLMs to wind turbine maintenance logs, it also highlights several promising avenues for future research. Further development in these areas could transition this methodology from a powerful labelling assistant into a fully integrated component of modern, data-driven O\&M strategies, ultimately enhancing reliability analysis and contributing to LCOE reduction. Key directions for future work include:
\begin{itemize}
    \item \textbf{Expanding the Benchmark:} The framework should be used to evaluate larger and more capable open-source models with better computational resources as well as other promising proprietary models, providing a continuously updated understanding of the state-of-the-art.
    
    \item \textbf{Domain-Specific Fine-Tuning:} A crucial next step is to fine-tune a high-performing open-source model on a large, expert-verified dataset. This would create a specialised, cost-effective tool tailored to the unique terminology of wind turbine maintenance, directly addressing the call for domain-specific models in the wind energy sector \cite{walgern_impact_2024}.
    
    \item \textbf{Large-Scale Data Processing:} For the task of labelling entire historical databases, research should investigate the performance and cost-effectiveness of batch processing API calls, which could unlock significant value from legacy data at a reduced cost.
    
    \item \textbf{Integration with Data Fusion:} The structured text output from this framework should be integrated with data fusion methodologies, such as probabilistic record linkage \cite{cox_enrichment_2022}, to connect maintenance events with synchronous SCADA data. This would create a truly comprehensive "Enriched Health History" for advanced root cause analysis.
    
    \item \textbf{Developing Advisory Systems:} The enriched dataset can serve as a knowledge base for training next-generation advisory agents. Future work could explore using this data to ground LLMs that recommend maintenance actions, incorporating essential safety-monitoring guardrails to ensure their reliability in a critical industrial setting \cite{walker_safellm_2024}.
    
    \item \textbf{Quantifying Business Impact:} Finally, a longitudinal study should be conducted in an operational setting to quantify the tangible business impact of an LLM-assisted Human-in-the-Loop workflow, measuring its effect on key reliability metrics, maintenance costs, and the overall LCOE.
\end{itemize}

\vspace{1cm}

\addcontentsline{toc}{section}{Acknowledgements}
\noindent\textbf{\large{Acknowledgements}}
\vspace{0.5cm}

\noindent This study is part of an ongoing PhD project on wind turbine reliability funded by the School of Engineering at the University of Edinburgh. Special acknowledgement is extended to Nadara, acting as the industrial partner, who provided the maintenance logs for this analysis. The authors also thank EDINA for providing access to OpenAI's API keys used in the study.

\vspace{1cm}

\addcontentsline{toc}{section}{Data and Code Availability}
\noindent\textbf{\large{Data and Code Availability}}
\vspace{0.5cm}

\noindent To ensure the reproducibility and transparency of this research, the complete source code, configuration files, and supplementary materials have been made publicly available in a GitHub repository, which can be accessed on our \textcolor{blue}{\href{https://github.com/mvmalyi/wind-farm-maintenance-logs-labelling-with-llms}{GitHub Repository}}. Please note that due to its commercially sensitive nature, the maintenance log dataset used in this study cannot be shared. However, to demonstrate the framework's functionality, the repository includes a sample dataset with a similar structure, allowing users to replicate the workflow. The open-source code is intended for users to apply to their own proprietary data, either for benchmarking LLMs or for data labelling.

\cleardoublepage
\phantomsection
\addcontentsline{toc}{section}{References}
\bibliographystyle{IEEEtran} 
\bibliography{maint-logs-llms} 

\begin{thebibliography}{10}
\providecommand{\url}[1]{#1}
\csname url@samestyle\endcsname
\providecommand{\newblock}{\relax}
\providecommand{\bibinfo}[2]{#2}
\providecommand{\BIBentrySTDinterwordspacing}{\spaceskip=0pt\relax}
\providecommand{\BIBentryALTinterwordstretchfactor}{4}
\providecommand{\BIBentryALTinterwordspacing}{\spaceskip=\fontdimen2\font plus
\BIBentryALTinterwordstretchfactor\fontdimen3\font minus \fontdimen4\font\relax}
\providecommand{\BIBforeignlanguage}[2]{{%
\expandafter\ifx\csname l@#1\endcsname\relax
\typeout{** WARNING: IEEEtran.bst: No hyphenation pattern has been}%
\typeout{** loaded for the language `#1'. Using the pattern for}%
\typeout{** the default language instead.}%
\else
\language=\csname l@#1\endcsname
\fi
#2}}
\providecommand{\BIBdecl}{\relax}
\BIBdecl

\bibitem{hodkiewicz_cleaning_2016}
M.~Hodkiewicz and M.~T.-W. Ho, ``Cleaning historical maintenance work order data for reliability analysis,'' \emph{Journal of Quality in Maintenance Engineering}, vol.~22, no.~2, pp. 146--163, May 2016.

\bibitem{hahn_recommended_2017}
B.~Hahn, T.~Welte, S.~Faulstich, P.~Bangalore, C.~Boussion, K.~Harrison, E.~{Miguelanez-Martin}, F.~O'Connor, L.~Pettersson, C.~Soraghan, C.~{Stock-Williams}, J.~Dalsgaard~S{\o}rensen, G.~Van~Bussel, and J.~Vatn, ``Recommended practices for wind farm data collection and reliability assessment for {{O}}\&{{M}} optimization,'' \emph{Energy Procedia}, vol. 137, pp. 358--365, 2017.

\bibitem{salo_work_2019}
E.~Salo, D.~McMillan, and R.~Connor, ``Work {{Orders}} - {{Value}} from {{Structureless Text}} in the {{Era}} of {{Digitisation}},'' in \emph{{{SPE Offshore Europe Conference}} and {{Exhibition}}}.\hskip 1em plus 0.5em minus 0.4em\relax Aberdeen, UK: SPE, Sep. 2019.

\bibitem{lutz_digitalization_2022}
M.-A. Lutz, J.~Walgern, K.~Beckh, J.~Schneider, S.~Faulstich, and S.~Pfaffel, ``Digitalization {{Workflow}} for {{Automated Structuring}} and {{Standardization}} of {{Maintenance Information}} of {{Wind Turbines}} into {{Domain Standard}} as a {{Basis}} for {{Reliability KPI Calculation}},'' \emph{Journal of Physics: Conference Series}, vol. 2257, no.~1, p. 012004, Apr. 2022.

\bibitem{salo_analysis_2017}
E.~Salo, ``Analysis of {{SAP}} work order data by turbine technology type for onshore wind,'' Master's thesis, University of Strathclyde, Glasgow, UK, 2017.

\bibitem{lutz_kpi_2023}
M.-A. Lutz, B.~Sch{\"a}fermeier, R.~Sexton, M.~Sharp, A.~Dima, S.~Faulstich, and J.~M. Aluri, ``{{KPI Extraction}} from {{Maintenance Work Orders}}---{{A Comparison}} of {{Expert Labeling}}, {{Text Classification}} and {{AI-Assisted Tagging}} for {{Computing Failure Rates}} of {{Wind Turbines}},'' \emph{Energies}, vol.~16, no.~24, p. 7937, Dec. 2023.

\bibitem{walgern_impact_2024}
J.~Walgern, K.~Beckh, N.~Hannes, M.~Horn, M.-A. Lutz, K.~Fischer, and A.~Kolios, ``Impact of using text classifiers for standardising maintenance data of wind turbines on reliability calculations,'' \emph{IET Renewable Power Generation}, vol.~18, no.~15, pp. 3463--3479, Nov. 2024.

\bibitem{walshe_automatic_2025}
T.~Walshe, S.~Y. Moon, C.~Xiao, Y.~Gunawardana, and F.~Silavong, ``Automatic {{Labelling}} with {{Open-source LLMs}} using {{Dynamic Label Schema Integration}},'' \emph{arXiv}, no. arXiv:2501.12332, Jan. 2025.

\bibitem{walker_safellm_2024}
C.~Walker, C.~Rothon, K.~Aslansefat, Y.~Papadopoulos, and N.~Dethlefs, ``{{SafeLLM}}: {{Domain-Specific Safety Monitoring}} for {{Large Language Models}}: {{A Case Study}} of {{Offshore Wind Maintenance}},'' \emph{arXiv}, no. arXiv:2410.10852, Oct. 2024.

\bibitem{cox_enrichment_2022}
R.~Cox, ``Enrichment of {{Wind Turbine Health History}} for {{Condition-Based Maintenance}},'' Ph.D. dissertation, Durham University, Durham, UK, 2022.

\end{thebibliography}

\end{document}